\documentclass[11pt,a4paper]{article}
\usepackage[hyperref]{acl2020}

\usepackage{pifont}
\usepackage{times}
\usepackage{latexsym}
\usepackage[colorinlistoftodos,prependcaption]{todonotes}
\usepackage{tabulary,booktabs}
\usepackage{float}
\usepackage{tablefootnote}
\usepackage{amsfonts}
\usepackage{color,soul}
\usepackage{multirow}
\usepackage{amsmath}
\usepackage{colortbl}
\usepackage{color, xcolor}
\usepackage{bm}
\usepackage[subtle]{savetrees}
\usepackage{float}
\usepackage{todonotes}

\newcommand{\softmax}{\sigma}
\newcommand{\sigmoid}{\phi}
\usepackage{url}

\usepackage{microtype}

\aclfinalcopy 

\title{End-to-End Bias Mitigation by Modelling Biases in Corpora}

\author{Rabeeh Karimi Mahabadi \\
  EPFL, Switzerland \\
  Idiap Research Institute, \\Switzerland \\
  \texttt{rabeeh.karimi@idiap.ch} \\\And
  Yonatan Belinkov \\
  Harvard University and \\ 
Massachusetts Institute of Technology,\\
Cambridge, MA, USA \\ 
  \texttt{\hspace{3em} belinkov@seas.harvard.edu} \\\And
  James Henderson \\
 Idiap Research Institute,\\
  Switzerland \\
  \texttt{james.henderson@idiap.ch} \\
  }

\date{}

\begin{document}
\maketitle

\begin{abstract}
Several recent studies have shown that strong natural language understanding (NLU) models are prone to relying on  unwanted dataset \emph{biases} without learning the underlying task,
resulting in models that fail to generalize to out-of-domain datasets and are likely to perform poorly in real-world scenarios. We propose two learning strategies to train neural models, which
are more robust to such biases and transfer better to out-of-domain datasets. 
The biases are specified in terms of one or more \emph{bias-only models}, which learn to leverage the dataset biases.
During training, the bias-only models' predictions are used to adjust the loss of the base model to reduce its reliance on biases by down-weighting the biased examples and focusing the training on the \emph{hard} examples.
We experiment on large-scale natural language inference and fact verification benchmarks, evaluating on out-of-domain datasets that are specifically designed to assess the robustness of models against known biases in the training data. Results show that our debiasing methods greatly improve robustness in all settings and better transfer to other textual entailment datasets. Our code and data are publicly available in \url{https://github.com/rabeehk/robust-nli}.
\end{abstract}

  \begin{figure*}[h]
 \centering 
    \includegraphics[width=0.96\textwidth]{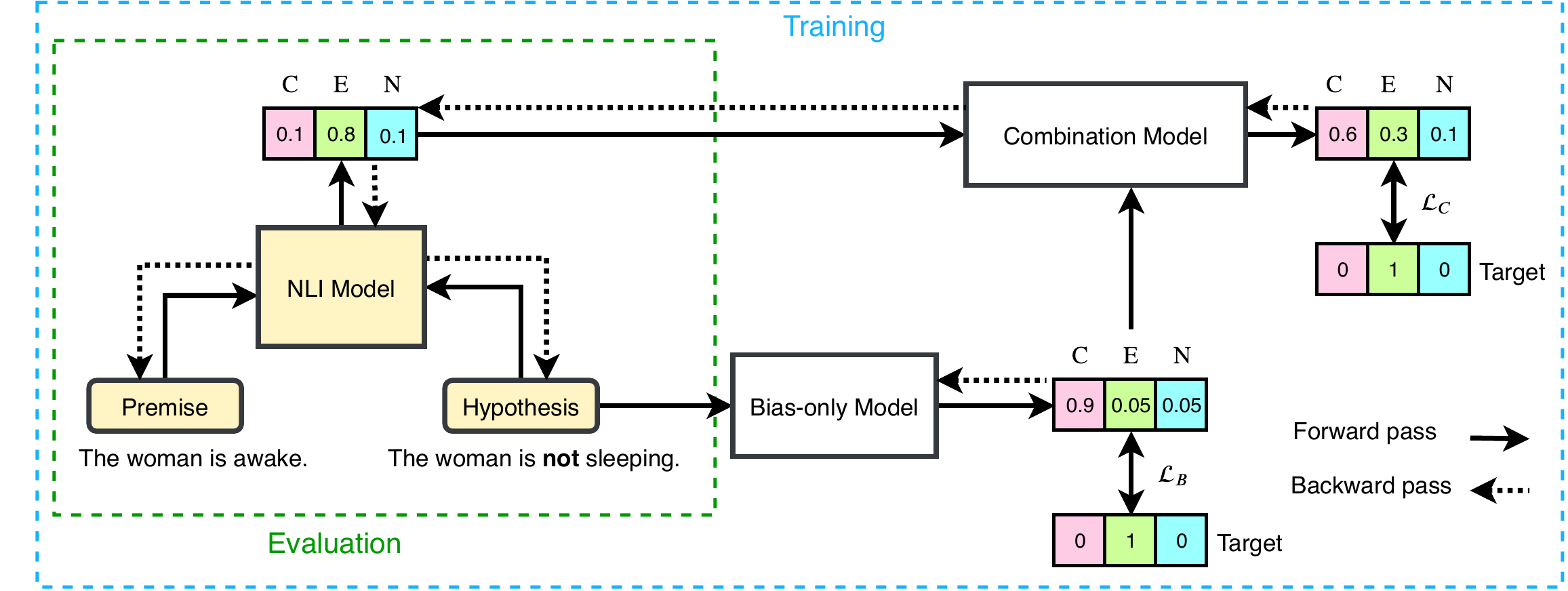}
    \caption{An illustration of our debiasing strategies applied to an NLI model. 
    The bias-only model only sees the hypothesis,  where negation words like ``not'' are highly correlated with the contradiction label. We train a robust NLI model by training it in combination with the bias-only model and motivate it to learn different strategies than the ones used in the bias-only model. The robust NLI model does not rely on the shortcuts and obtains improved performance on the test set.} 
    \label{fig:model}
\end{figure*}

\section{Introduction}

Recent neural models \citep{devlin2018bert, radford2018improving, chen2017enhanced} have achieved high and even near human-performance on several large-scale natural
language understanding benchmarks.  However, it has been demonstrated that neural models tend to rely on existing idiosyncratic biases in the datasets,
and leverage superficial correlations between the label and existing shortcuts in the training dataset to perform surprisingly well,\footnote{We use biases, heuristics or shortcuts interchangeably.} 
without learning the underlying task \citep{kaushik2018much, gururangan2018annotation, poliak2018hypothesis, schuster2019towards,  mccoy-etal-2019-right}. For instance, natural language
inference (NLI) is supposed to test the ability of a model to determine whether a hypothesis sentence (\emph{There is no teacher in the room}) can be inferred from a premise sentence (\emph{Kids work at computers with a teacher's help}) \citep{dagan2006pascal}.\footnote{The 
given sentences are in the contradictory relation, and the hypothesis cannot be inferred from the premise.}  However, recent work has demonstrated that large-scale NLI benchmarks contain annotation
artifacts; certain words in the hypothesis that are highly indicative of inference class  and allow models that do not consider the premise to perform unexpectedly well \citep{poliak2018hypothesis, gururangan2018annotation}. As an example, in some 
NLI benchmarks, negation words such as ``nobody'', ``no'', and ``not'' in the hypothesis are often highly correlated with the contradiction label.

As a result of the existence of such biases, models
exploiting statistical shortcuts during training often perform poorly on out-of-domain datasets, especially if the datasets are carefully designed to limit the spurious cues. 
To allow proper evaluation, recent studies have tried to create new evaluation datasets that do not contain such biases \citep{gururangan2018annotation, schuster2019towards, mccoy-etal-2019-right}. Unfortunately, it is hard to avoid spurious statistical
cues in the construction of large-scale benchmarks, and collecting new datasets is costly \citep{sharma2018tackling}. It is, therefore, crucial to develop techniques to reduce the reliance on biases during the training of the neural models.

We propose two end-to-end debiasing techniques that can be used when the existing bias patterns are identified. These methods work by adjusting the cross-entropy loss to reduce the biases learned from the training dataset, down-weighting the biased examples so that the model focuses on learning the hard examples. 
Figure~\ref{fig:model} illustrates an example of applying our strategy to prevent an NLI model from predicting
the labels using existing biases in the hypotheses, where the bias-only model only sees the hypothesis.
Our strategy involves adding this bias-only branch $f_B$ on top of the base model $f_M$ during training.
We then compute the combination of the two models $f_C$ in a way that motivates the base model to learn different 
strategies than the ones used by the bias-only branch $f_B$. At the end of the training, we remove the bias-only
classifier and use the predictions of the base model.

In our first proposed method, Product of Experts, the training loss is computed on an ensemble of the base model and the bias-only model, which reduces the base model's loss for the examples that the bias-only model classifies correctly.  For the second method, Debiased Focal Loss, the bias-only predictions
are used to directly weight the loss of the base model, explicitly modulating the loss depending on the accuracy of the bias-only model.
We also extend these methods to be robust against multiple sources of bias by training multiple bias-only models.

Our approaches are simple and highly effective. They require training only a simple model on top of the base model. They are model agnostic and 
general enough to be applicable for addressing common biases seen in many datasets in different domains. 

We evaluate our models on challenging benchmarks in textual entailment and fact verification, including HANS (Heuristic Analysis for NLI Systems) \citep{mccoy-etal-2019-right}, hard NLI sets  \citep{gururangan2018annotation} of Stanford Natural Language Inference (SNLI) \citep{bowman2015large} and MultiNLI (MNLI) \citep{williams2018broad}, and FEVER Symmetric test set \citep{schuster2019towards}. The selected datasets are highly challenging and have been carefully designed to be unbiased to allow proper evaluation of the out-of-domain performance of the models. We additionally construct hard MNLI datasets from MNLI development sets to facilitate the out-of-domain evaluation on this dataset.\footnote{Removing the need to submit to an online evaluation system for MNLI hard test sets.} 
 We show that including our strategies on training baseline models, including BERT \citep{devlin2018bert}, provides a substantial gain on out-of-domain performance in all the experiments.

In summary, we make the following contributions:
1) Proposing two debiasing strategies to train  neural models robust to  dataset bias. 
2) An empirical evaluation of the methods on two large-scale NLI datasets  and  a fact verification benchmark; obtaining a substantial gain on their challenging out-of-domain data, including 7.4 points on HANS, 4.8 points on SNLI hard set, and 9.8 points  on  FEVER symmetric test set, setting a new state-of-the-art. 
3) Proposing debiasing strategies capable of combating multiple sources of bias. 4) Evaluating the transfer performance of the  debiased models on 12 NLI datasets and demonstrating improved transfer to other NLI benchmarks. To facilitate future work, we release our datasets and code.

\section{Related Work}

To address dataset biases, researchers have proposed to augment  datasets by balancing the existing cues~\citep{schuster2019towards} or to create an adversarial dataset~\citep{jia2017adversarial}. However, collecting new datasets, especially at a large scale, is costly, and thus remains an unsatisfactory solution. It is,  therefore,  crucial to develop strategies to allow models to be trained on the existing biased datasets. 

\citet{schuster2019towards} propose to  first compute the n-grams in the dataset's claims that are the most associated with each fact-verification label. They then solve an optimization problem to assign a balancing weight to each training sample to  alleviate the biases. In contrast, we propose several end-to-end debiasing strategies.  Additionally, \citet{belinkov-etal-2019-dont}  propose adversarial techniques  to remove from the NLI sentence encoder the features that allow a
hypothesis-only model to succeed. However, we believe that in general,
the features used by the hypothesis-only model can include some
information necessary to perform the NLI task, and removing such
information from the sentence representation can hurt the performance
of the full model.  Their approach consequently degrades the performance on the hard SNLI
 set, which is expected to be less biased.  In contrast, we propose to train a bias-only model to use its predictions to
  dynamically  adapt the classification loss to reduce the importance of the most biased examples. 

Concurrently to our work,~\citet{clark2019dont} and \citet{he-etal-2019-unlearn} have also proposed to use the product of experts (PoE) models for avoiding biases. They train their models in two stages, first training a bias-only model and then using it to train a robust model. In contrast, our methods  are trained in an end-to-end manner, which is convenient in practice. We additionally show that our proposed Debiased Focal Loss model is an effective method to reduce biases, sometimes superior to PoE. We have evaluated on new domains of NLI hard sets and fact verification.  Moreover, we have included an analysis showing that our debiased models indeed have lower correlations with the bias-only models, and have extended our methods to guard against multiple bias patterns simultaneously. We furthermore study transfer performance to other NLI datasets.

\section{Reducing Biases}
\paragraph{Problem formulation} We consider a general multi-class classification problem. Given a dataset $\mathcal{D} = \{\bm{x_i}, y_i\}_{i=1}^N$ consisting of
the input data $\bm{x_i} \in \mathcal{X}$, and labels $y_i \in \mathcal{Y}$, the goal of the base model 
is to learn a mapping $f_M$ parameterized by $\theta_M$ that computes the predictions over the label space given the input data, shown as  $f_M: \mathcal{X} \rightarrow \mathbb{R}^{|\mathcal{Y}|}$. 
Our goal is to optimize $\theta_M$ parameters such that we build a model that is more 
resistant to benchmark dataset biases, to improve its robustness to domain changes where the biases typically observed in the training data do not exist  in the evaluation dataset.

The key idea of our approach, depicted in Figure~\ref{fig:model}, is first to identify the dataset biases that the base model is 
susceptible to relying on, and define a bias-only model to capture them. We then propose two strategies to incorporate this bias-only knowledge into the 
training of the base model to make it robust against the biases.  After training, we remove the bias-only model and use the predictions of the base model. 

\subsection{Bias-only Branch} 
We assume that we do not have access to any data from the out-of-domain dataset, so we need to know a priori about the possible types of shortcuts 
 we would like the base model to avoid relying on. Once these patterns are identified, we train a bias-only model designed to 
capture the identified shortcuts that only uses \emph{biased features}. For instance, a hypothesis-only model in the large-scale
NLI datasets can correctly classify the majority of samples 
using annotation  artifacts \citep{poliak2018hypothesis, gururangan2018annotation}. Motivated by this work, our bias-only model for
NLI only uses hypothesis sentences. Note that the bias-only model can, in general, have any form,
and is not limited to models using only a part of the input data. For instance, on the HANS dataset, our bias-only model makes use of syntactic heuristics and similarity features (see Section~\ref{sec:syntactic_bias}). 

Let $\bm{x_i^b} \in \mathcal{X}^b$ be \emph{biased features} of $\bm{x_i}$ that are predictive of $y_i$. We then formalize this bias-only model as a mapping
$f_B:  \mathcal{X}^b \rightarrow \mathbb{R}^{|\mathcal{Y}|}$, parameterized by $\theta_B$ and trained using cross-entropy (CE) loss $\mathcal{L}_B$:
\begin{align}
\mathcal{L}_B(\theta_B) = -\frac{1}{N} \sum_{i=1}^N \log(\softmax(f_B^{y_i}(\bm{x_i^b}; \theta_B))), \label{eq:bias-cross-entropy} 
\end{align}
where $f_B^j(\bm{x_i^b}, \theta_B)$ is the $j$th element of $f_B(.)$, and $\sigma(u^j)=e^{u^j}/\sum_{k=1}^{|\mathcal{Y}|} e^{u^k}$ is the softmax function.

\subsection{Proposed Debiasing Strategies}\label{sec:debiasing-methods}

We propose two strategies to incorporate the bias-only $f_B$ knowledge into the training of the base model $f_M$. In our strategies, the predictions of the bias-only model are combined with either the predictions of the base model or its error, to down-weight the loss for the examples that the bias-only model can predict correctly. We then update parameters of the base model $\theta_M$ based on this modified loss $\mathcal{L}_C$. Our learning strategies are end-to-end. Therefore, to prevent the base model from learning the biases, the bias-only loss $\mathcal{L}_B$ is not back-propagated to any shared parameters of the base model, such as a shared 
sentence encoder.

\subsubsection{Method 1: Product of Experts} \label{sec:product_of_experts}
Our first approach is based on the \emph{product of experts} (PoE) method~\citep{hinton2002training}. 
Here, we use this method to combine the bias-only and base
model's predictions by computing the element-wise product $\odot$ between their predictions as $\softmax( f_B(\bm{x_i^b})) \odot  \softmax(f_M(\bm{x_i}))$. We compute this combination
in the logarithmic space, making it appropriate for the normalized exponential below:
 \begin{align}
f_C(\bm{x_i}, \bm{x_i^b})=\log(\softmax(f_B(\bm{x_i^b})))+\log(\softmax(f_M(\bm{x_i}))), \nonumber 
\end{align}
The key intuition behind this model is to combine the probability distributions of the bias-only and the base model to allow them to make predictions based on different
characteristics of the input; the bias-only branch covers prediction based on biases, and the base model focuses on learning the actual task. Then the base model parameters $\theta_M$ are trained using the cross-entropy loss $\mathcal{L}_C$ of the combined classifier $f_C$: 
\begin{align}
\mathcal{L}_C(\theta_M;\theta_B) = -\frac{1}{N} \sum_{i=1}^N  \log(\softmax(f_C^{y_i}(\bm{x_i}, \bm{x_i^b}))).
\label{eq:ensemble-cross-entropy} 
\end{align}
When updating the base model parameters  using this loss, the predictions of the bias-only model decrease the updates for examples that it can accurately predict. 

\paragraph{Justification:}  Probability of label $y_i$ for the example $\bm{x_i}$ in the PoE model is computed as: 
\begin{align}
\softmax(f_C^{y_i}(\bm{x_i}, \bm{x_i^b})) = \frac{\softmax(f_B^{y_i}(\bm{x_i^b}))\softmax(f_M^{y_i}(\bm{x_i}))}{\sum_{k=1}^{|\mathcal{Y}|} \softmax(f_B^k(\bm{x_i^b}))\softmax(f_M^k(\bm{x_i}))} \nonumber
\end{align}
Then the gradient of cross-entropy loss of the combined classifier~\eqref{eq:ensemble-cross-entropy} 
w.r.t $\theta_M$ is~\citep{hinton2002training}:
\begin{align}
&\nabla_{\theta_M} \mathcal{L}_C(\theta_M;\theta_B) =  -\frac{1}{N} \sum_{i=1}^N \sum_{k=1}^{|\mathcal{Y}|} \bigg[ 
\nonumber \\
&
\left( \delta_{y_i k} - \softmax(f_C^k(\bm{x_i}, \bm{x_i^b})) \right)
\nabla_{\theta_M}\log(\softmax(f_M^k(\bm{x_i})))
\bigg],\nonumber
\end{align}
where $\delta_{y_ik}$ is 1 when $k{=}y_i$ and 0 otherwise. Generally, the closer the ensemble's prediction $\softmax(f_C^k(.))$ is to the target $\delta_{y_ik}$, the more the gradient is decreased through the modulating term, which only happens when the bias-only and base models are both capturing biases.

In the extreme case, when the bias-only model correctly classifies the sample, $\softmax(f_C^{y_i}(\bm{x_i}, \bm{x_i^b}))=1$ and therefore  $\nabla_{\theta_M}\mathcal{L}_C(\theta_M; \theta_B)=0$, the biased
examples are ignored during training. Conversely, when the example is fully unbiased, the bias-only classifier predicts the uniform distribution over all labels $\softmax(f_B^{k}(\bm{x_i^b}))=\frac{1}{|\mathcal{Y}|}$ for $k \in \mathcal{Y}$,
therefore $\softmax(f_C^{y_i}(\bm{x_i}, \bm{x_i^b}))=\softmax(f_M^{y_i}(\bm{x_i}))$ and the gradient of ensemble classifier remains the same as the CE loss.

\subsubsection{Method 2: Debiased Focal Loss} \label{sec:debiased_focal_loss}
Focal loss was originally proposed in \citet{lin2017focal} to improve a single classifier by down-weighting the well-classified points. We propose a novel variant of this loss that leverages
the bias-only branch's predictions to reduce the relative importance of the most
biased examples and allows the model to focus on learning the \emph{hard} examples. We define
\emph{Debiased Focal Loss} (DFL) as:
\begin{align}
&\mathcal{L}_{C}(\theta_M; \theta_B) =  \label{eq:debiased_focal_loss}\\
&-\frac{1}{N} \sum_{i=1}^N  \left(1-\softmax(f_B^{y_i}(\bm{x_i^b}))\right)^{\gamma} \nonumber  \log(\softmax(f_M^{y_i}(\bm{x_i})))
\end{align}
where $\gamma$ is the focusing parameter, which impacts the down-weighting rate. When $\gamma$ is set to 0, DFL is 
equivalent to the cross-entropy loss. For $\gamma > 0$, as the value of $\gamma$ is increased, the effect of down-weighting is increased. We set $\gamma=2$ through all experiments, which works well in practice, and avoid fine-tuning it further. We note the properties of this loss: (1) When the example $\bm{x_i}$ is unbiased, and the bias-only branch does not do
well, $\softmax(f_B^{y_i}(\bm{x_i^b}))$ is small, therefore the scaling factor is close to $1$, and the loss remains unaffected. (2) As the sample is more biased and $\softmax(f_B^{y_i}(\bm{x_i^b}))$ is closer to 1, the modulating factor approaches 0 and the loss for the most biased examples is down-weighted. 

\subsection{RUBi baseline \citep{cadene2019rubi}}\label{sec:rubi_methods}
We compare our models to RUBi~\citep{cadene2019rubi}, a recently proposed model to 
alleviate unimodal biases learned by Visual Question Answering (VQA) models. \citet{cadene2019rubi}'s study is limited to VQA datasets. 
We, however, evaluate the effectiveness of their formulation on
 multiple challenging NLU benchmarks.
RUBi consists in first applying a sigmoid function $\phi$ to the bias-only model's predictions to obtain a mask containing an importance weight between 0 and 1 for each label. It then computes the element-wise product between the obtained mask and the base model's predictions:
\begin{align}
f_C(\bm{x_i}, \bm{x_i^b}) = f_M(\bm{x_i}) \odot \sigmoid (f_B(\bm{x_i^b})), \nonumber 
\end{align}
The main intuition is to dynamically adjust the predictions of the base model to prevent it from leveraging the shortcuts.
Then the parameters of the base model $\theta_M$ are updated by back-propagating the
cross-entropy loss $\mathcal{L}_C$ of the combined classifier.

\subsection{Joint Debiasing Strategies}
Neural models can, in practice, be prone to multiple types of biases in the datasets. We, therefore, propose  methods for combining several bias-only models. To avoid learning relations between biased features, we do not consider training a classifier on top of their concatenation.

Instead, let $\{\bm{x_i^{b_j}}\}_{j=1}^K$ be different sets of \emph{biased features} of $\bm{x_i}$ that are predictive of $y_i$, and let $f_{B_j}$ be an individual bias-only model capturing $\bm{x_i^{b_j}}$. Next, we extend our debiasing strategies to handle multiple bias patterns.

\paragraph{Method 1: Joint Product of Experts}
We extend our proposed PoE model to multiple bias-only models by computing the element-wise product between the predictions of bias-only models and the base model as: $\sigma(f_{B_1}(\bm{x_i^{b_1}}))\odot \dots \odot \sigma(f_{B_K}(\bm{x_i^{b_K}})) \odot \sigma(f_M(\bm{x_i}))$, computed in the logarithmic space:
\begin{align}
f_C(\bm{x_i}, \{\bm{x_i^{b_j}}\}_{j=1}^K)&=\sum_{j=1}^K\log(\softmax(f_{B_j}(\bm{x_i^{b_j}})))\nonumber\\
&+\log(\softmax(f_M(\bm{x_i}))). \nonumber 
\end{align}
Then the base model parameters $\theta_M$ are trained using the cross-entropy loss of the combined classifier $f_C$.

\paragraph{Method 2: Joint Debiased Focal Loss}
To extend DFL to handle multiple bias patterns, we first compute the element-wise average of the predictions of the multiple bias-only models: $f_B(\{\bm{x_i^{b_j}}\}_{j=1}^K) = \frac{1}{K}\sum_{j=1}^K f_{B_j}(\bm{\bm{x_i^{b_j}}})$, and then compute the DFL~\eqref{eq:debiased_focal_loss} using the computed joint bias-only model.

\section{Evaluation on Unbiased Datasets}\label{sec:experiments_unbiased}
We provide experiments on a fact verification (FEVER) and two large-scale NLI datasets (SNLI and MNLI).  We evaluate
the models' performance on recently-proposed challenging unbiased evaluation sets. We use the BERT~\citep{devlin2018bert} implementation of~\citet{Wolf2019HuggingFacesTS} as our main baseline, known to work well for these tasks.  In all the experiments, we use the default hyperparameters 
of the baselines.

\subsection{Fact Verification} \label{sec:fact_experiments}
\paragraph{Dataset:} The FEVER dataset contains claim-evidence pairs generated from Wikipedia. \citet{schuster2019towards} collected a new evaluation set for the FEVER dataset to avoid the idiosyncrasies
observed in the claims of this benchmark. They made the original claim-evidence pairs of the FEVER  evaluation dataset
symmetric, by augmenting them and making each claim and evidence appear with each label.
Therefore, by balancing the artifacts, relying on statistical cues in claims to classify 
samples is equivalent to a random guess. The collected dataset  is challenging, and the performance
of the models relying on biases evaluated on this dataset drops significantly. 

\paragraph{Base models:} We consider  BERT as the base model, which works the best on this dataset \citep{schuster2019towards}, and 
predicts the relations based on the concatenation of the claim and the evidence with a delimiter token (see Appendix~\ref{sec:appendix_verification}).

\paragraph{Bias-only model:} The bias-only model predicts the labels using only claims as input.

\paragraph{Results:} Table~\ref{tab:verification_results} shows the results. Our proposed debiasing methods, PoE and DFL, are highly effective,
boosting the performance of the baseline by 9.8 and 7.5  points respectively, significantly surpassing
the prior work of~\citet{schuster2019towards}. 
\begin{table}[H]
\centering 
    \begin{tabular}{@{~}l@{~}l@{~~~~}l@{}@{~~~}l} 
        \toprule 
        \textbf{Loss}     & \textbf{Dev} & \textbf{Test} & \bm{$\Delta$}\\  
        \toprule 
        CE               & 85.99  & 56.49 &  \\ 
         RUBi     & 86.23 &  57.60 & +1.1 \\
        \citet{schuster2019towards} 
        & 84.6 & \textbf{61.6} & \textbf{+5.1} \\
        \midrule 
        DFL         & 83.07 &  64.02 & +7.5 \\
        PoE & 86.46 &  \textbf{66.25} & \textbf{+9.8} \\
        \bottomrule 
    \end{tabular} \vspace{-0.4em}
    \caption{Results on FEVER development and symmetric test set. $\bm{\Delta}$ are absolute differences with CE loss.\nolinebreak}
    \label{tab:verification_results}
\end{table}

\subsection{Natural Language Inference}   \label{sec:entailment_results}
\paragraph{Datasets:} We evaluate on hard datasets of SNLI and MNLI \citep{gururangan2018annotation}, which are the splits of these datasets
where a hypothesis-only model cannot correctly predict the labels.
\citet{gururangan2018annotation} show that the success of the recent
textual entailment models is attributed to the \emph{biased} examples,
and the performance of these models is substantially lower on the \emph{hard} sets. 

\paragraph{Base models:} We consider BERT and InferSent~\citep{conneau2017supervised} as our base models. We choose InferSent to be able to compare with 
the prior work of~\citet{belinkov2019adversarial}.

\paragraph{Bias-only model:} The bias-only model predicts the labels using the hypothesis (Appendix~\ref{appendix:entailment}).

\paragraph{Results on SNLI:} Table~\ref{tab:snli_hard_test_results} shows the SNLI results. With InferSent, DFL and PoE result in 4.1 and 4.8 points gain. With BERT, DFL
and PoE improve the results by 2.5 and 1.6 absolute points. Compared to the prior work of \citet{belinkov2019adversarial} (AdvCls), our
PoE model obtains a 7.4 points gain, setting a new state-of-the-art.   

\begin{table}[H]
    \centering
    \resizebox{0.5\textwidth}{!}{
          \begin{tabular}{lllllllllllll} 
        \toprule 
        \multirow{2}{*}{\textbf{Loss}} & \multicolumn{3}{c}{\textbf{BERT}}  & \multicolumn{3}{c}{\textbf{InferSent}} \\ 
          \cmidrule(r){2-4} \cmidrule(l){5-7} 
                 &  \textbf{Test} & \textbf{Hard} & \bm{$\Delta$} & \textbf{Test} & \textbf{Hard} & \bm{$\Delta$}\\  
        \toprule 
        CE        & 90.53   & 80.53 &&84.24 &68.91&\\
        RUBi        & 90.69   & \textbf{80.62} &\textbf{+0.1} &83.93 &\textbf{69.64}&\textbf{+0.7}\\
        AdvCls*  & --- & --- & --- &83.56  & {66.27} &{-2.6} \\ 
        AdvDat* & --- & --- & --- &78.30  & 55.60  &-13.3\\
        \midrule 
        DFL        & 89.57  & \textbf{83.01}&\textbf{+2.5} & 73.54 & 73.05&+4.1\\ 
        PoE & 90.11  & 82.15& +1.6& 80.35 & \textbf{73.69}&\textbf{+4.8}\\ 
        \bottomrule 
    \end{tabular}}
     \caption{Results on the SNLI test, hard set, and differences with CE loss. *: results from~\citet{belinkov2019adversarial}.}
    \label{tab:snli_hard_test_results}
\end{table}

\paragraph{Results on MNLI:} 
We construct hard sets from the validation sets  of MNLI Matched and  Mismatched (MNLI-M). Following \citet{gururangan2018annotation}, we train a \texttt{fastText} classifier \citep{joulin2017bag} that
predicts the labels using only the hypothesis  and consider the subset on which it fails as hard examples.

We report the results on MNLI mismatched  sets in Table~\ref{tab:mnli_hard_test_results_all} (see Appendix~\ref{appendix:entailment} for similar results on MNLI matched). With BERT,  DFL and PoE obtain 1.4 and 1.7 points gain on the hard development set, while with InferSent,  they improve the results by 2.5 and 2.6 points.  To comply with limited access to the MNLI submission system, we evaluate only the best result of the baselines and 
our models on the test sets. Our PoE
model improves the performance on the hard test set by 1.1 points  while retaining in-domain accuracy. 

\begin{table}[H]
    \centering
    \resizebox{0.5\textwidth}{!}{
    \begin{tabular}{lllllllllllllll} 
        \toprule 
         & \multicolumn{3}{c}{\textbf{BERT}} & \multicolumn{3}{c}{\textbf{InferSent}} \\  
         \cmidrule(r){2-4} \cmidrule(l){5-7} 
        {\vspace{-0.75em} \textbf{Loss}} &  \multicolumn{2}{c}{} & \multicolumn{2}{c}{}\\ 
                 & \textbf{MNLI} & \textbf{Hard} & \bm{$\Delta$} &\textbf{MNLI} & \textbf{Hard} & \bm{$\Delta$} \\  
        \toprule 
         \multicolumn{7}{c}{\textbf{Development set results}}\\
         \toprule 
        CE &84.53  &77.55&& 69.99 & 56.53 &\\
        RUBi & 85.17 & 78.63& +1.1 & 70.53 & \textbf{58.08}&\textbf{+1.5}\\
        \midrule 
        DFL  & 84.85 & 78.92 &+1.4& 61.12 & 59.05&+2.5\\ 
        PoE &  84.85 & \textbf{79.23} &\textbf{+1.7}& 65.85& \textbf{59.14}&\textbf{+2.6}\\ 
    \bottomrule
    \multicolumn{7}{c}{\textbf{Test set results}} \\
    \toprule
    CE & 83.51 & 75.75 & & ---&--- &---  \\
    PoE & 83.47 & \textbf{76.83} &\textbf{+1.1} & ---&---&--- \\
    \bottomrule
    \end{tabular}}
 \caption{Results on MNLI mismatched benchmark and MNLI mismatched hard set. $\bm{\Delta}$ are absolute differences with CE loss.} 
    \label{tab:mnli_hard_test_results_all}
\end{table}

\subsection{Syntactic Bias in NLI}\label{sec:syntactic_bias}
\paragraph{Dataset:} \citet{mccoy-etal-2019-right} show that NLI models trained on MNLI can adopt superficial syntactic heuristics. They introduce HANS, consisting of several examples on which the syntactic heuristics fail. 

\paragraph{Base model:} We use BERT as our base model and train it on the MNLI dataset.

\paragraph{Bias-only model:} We consider the following features for the bias-only model. The first four features are based on the syntactic heuristics proposed in \citet{mccoy-etal-2019-right}: 
1) Whether all words in the hypothesis are included in the premise; 2) If the hypothesis is the contiguous subsequence of the premise; 3) If the hypothesis is a subtree in the
premise's parse tree; 4) The number of tokens shared between premise and hypothesis normalized by the number of tokens in the premise. We additionally include some 
similarity features: 5) The cosine similarity between premise and hypothesis's pooled token representations from BERT followed by min, mean, and max-pooling. We consider the same weight for contradiction and neutral labels in the bias-only loss to allow the model
to recognize entailment from not-entailment. During the evaluation, we map the neutral and contradiction labels to not-entailment.

\paragraph{Results:} \citet{mccoy2019berts} observe large variability in the linguistic generalization of neural models. We, therefore, report the averaged results across 4 runs with the standard deviation in Table~\ref{tab:hans_results}. PoE and DFL obtain 4.4 and 7.4 points gain (see Appendix~\ref{appendix:hans} for accuracy on individual heuristics of HANS). 

\begin{table}[H]
    \centering 
    \resizebox{0.5\textwidth}{!}{
    \begin{tabular}{llllllllllllll}
        \toprule 
        \textbf{Loss}    &  \textbf{MNLI} & \textbf{HANS}   & \bm{$\Delta$} \\ 
        \toprule 
        CE                  & 84.51 & 61.88 $\pm$1.9 &   \\ 
        RUBi                & 84.53 & 61.76$\pm$2.7  & -0.1 \\
        Reweight~\ding{163}    & 83.54 & \textbf{69.19} & \textbf{+7.3}\\  
        Learned-Mixin~\ding{163}  & 84.29 & 64.00 & +2.1\\
        Learned-Mixin+H~\ding{163}~\ding{68} & 83.97 & \textbf{66.15} & \textbf{+4.3} \\
        \midrule 
        PoE               & 84.19  &66.31$\pm$0.6  & +4.4\\ 
        DFL  &  83.95 & \textbf{69.26} $\pm$0.2 & \textbf{+7.4}\\
        DFL\ding{68}  &  82.76 & \textbf{71.95}$\pm$1.4 &  \textbf{+10.1}  \\
        \bottomrule 
    \end{tabular}}
    \vspace{-1ex}
     \caption{Results on MNLI Matched dev set and HANS. \ding{163}: results from~\citet{clark2019dont}. \ding{68}: perform hyper-parameter tuning. $\bm{\Delta}$ are differences with CE loss.}
    \label{tab:hans_results} 
\end{table}

We compare our results with the concurrent work of~\citeauthor{clark2019dont}, who propose a PoE model similar to ours, which gets similar results.  The main difference is that our models are trained end-to-end, which is convenient in practice, while  \citeauthor{clark2019dont}'s method requires two steps, first training a bias-only model and then using this pre-trained model to train a robust model. The Reweight baseline in \citeauthor{clark2019dont} is a special case of our DFL with $\gamma=1$ and performs similarly to our DFL method (using default $\gamma=2$). Their  Learned-Mixin+H method requires hyperparameter tuning. Since the assumption is not having access to any out-of-domain test data, and  there is no available dev set for HANS, it is challenging to perform hyper-parameter tuning. \citeauthor{clark2019dont} follow prior work~\citep{grand2019adversarial, ramakrishnan2018overcoming} and perform model section on the test set.

To provide a fair comparison, we consequently also tuned $\gamma$ in DFL by sweeping over $\{0.5, 1, 2, 3, 4\}$. DFL\ding{68} is the selected model, with $\gamma=3$.  With this hyperparameter tuning, DFL is even more effective, and our best result performs 2.8 points better than \citet{clark2019dont}.

\subsection{Jointly Debiasing Multiple Bias Patterns}
To evaluate combating multiple bias patterns, we jointly debias a base model on the hypothesis artifacts and syntactic biases. 
\paragraph{Base model:} We use BERT as our base model and train it on the MNLI dataset.
\begin{table}[H]
    \centering
      \resizebox{0.5\textwidth}{!}{
    \begin{tabular}{lllllllllllllll}
        \toprule 
        \textbf{Loss}     &  \textbf{MNLI}  & \textbf{Hard} & \bm{$\Delta$} &\textbf{HANS}   & \bm{$\Delta$}\\ 
        \toprule 
        CE                & 84.53  &  77.55 & & 61.88$\pm$1.9    & \\ 
        \midrule 
        PoE~\ding{168}    & 84.85 &     \textbf{79.23}  &\textbf{+1.7}&  60.43  &-1.5 \\
        DFL\ding{168}     & 84.85 &     78.92  &+1.4&  60.63  &-1.2\\
        \midrule 
        PoE~\ding{170}    & 84.55  &  77.90$\pm$0.3 & +0.4 &66.31$\pm$0.6  & +4.4  \\
        DFL\ding{170}     & 84.30  &  77.66$\pm$0.6 & +0.1 &\textbf{69.26}$\pm$0.2  & \textbf{+7.4}  \\
        \midrule 
        PoE-Joint         & 84.39  &  \textbf{78.61}$\pm$0.1 & \textbf{+1.1} & 68.04$\pm$1.2 &+6.2 \\
        DFL-Joint         & 84.49  &  78.36$\pm$0.4 & +0.8& \textbf{69.10}$\pm$0.7& \textbf{+7.2}\\
        \bottomrule 
    \end{tabular}}
     \caption{Results on MNLI mismatched dev set, MNLI mismatched hard set, and HANS when training independently to debias against either hypothesis artifacts (\ding{168}) or syntactic biases (\ding{170}), compared with jointly training to debias against both bias types. $\bm{\Delta}$: differences with baseline CE loss.}
    \label{tab:ensemble_results}
\end{table}

\paragraph{Bias-only models:} We use the hypothesis-only and syntactic bias-only models as in Sections \ref{sec:entailment_results} and ~\ref{sec:syntactic_bias}.
\paragraph{Results:}Table~\ref{tab:ensemble_results} shows the results. Models trained to be robust to hypothesis biases (\ding{168}) do not generalize to HANS. On the other hand,  models trained to be robust on HANS (\ding{170}) use a powerful bias-only model resulting in a slight improvement on MNLI mismatched hard dev set. We expect a slight degradation when debiasing for both biases since models need to select samples accommodating both debiasing needs. The jointly debiased models successfully obtain improvements on both datasets, which are close to the improvements on each dataset by the individually debiased  models.

\section{Transfer Performance}
To evaluate how well the baseline and proposed models generalize to solving textual entailment in domains that do not share the same annotation biases as the large NLI training sets, we take trained NLI models and test them on several NLI datasets.

\paragraph{Datasets:} We consider a total of 12 different NLI datasets.
We use the 11 datasets studied by \citet{poliak2018hypothesis}. These datasets include
MNLI, SNLI, SciTail \citep{khot2018scitail}, AddOneRTE (ADD1) \citep{pavlick2016most}, Johns Hopkins Ordinal Commonsense Inference (JOCI) \citep{zhang2017ordinal}, Multiple Premise Entailment (MPE) \citep{lai2017natural}, Sentences Involving Compositional Knowledge (SICK)~\citep{MARELLI14.363}, and three datasets from \citet{white2017inference} which are automatically generated from existing datasets for other NLP tasks including: Semantic Proto-Roles (SPR)~\citep{reisinger2015semantic}, Definite Pronoun Resolution (DPR)~\citep{rahman2012resolving}, FrameNet Plus (FN+) \citep{pavlick2015framenet+}, and the GLUE benchmark's diagnostic test~\citep{wang2018glue}. We additionally consider the Quora Question Pairs (QQP) dataset, where the task is to determine whether two given questions are semantically matching (duplicate) or not. 
As in \citet{gong2017natural}, we interpret duplicate question pairs as an entailment relation
and neutral otherwise.  We use the same split ratio mentioned by \citet{wang2017bilateral}.

Since the datasets considered have different label spaces, when evaluating on each target dataset, we map the model's labels to the corresponding target dataset's space. See Appendix~\ref{app:transfer} for more details.

We strictly refrained from using any out-of-domain data when evaluating on the unbiased split of the same benchmark in Section~\ref{sec:experiments_unbiased}. However, as shown by prior work~\citep{belinkov-etal-2019-dont}, since different NLI target datasets contain different amounts of the bias found in the large-scale NLI dataset, we need to adjust the amount of debiasing according to each target dataset. We consequently introduce a hyperparameter $\alpha$ for PoE to modulate the strength of the bias-only model in ensembling.  We follow prior work~\citep{belinkov-etal-2019-dont} and perform model selection on the dev set of each target dataset and then report results on the test set.\footnote{Since the test sets are not available for MNLI, we tune
on the matched dev set and evaluate on the mismatched dev
set or vice versa. For GLUE, we tune on MNLI mismatched dev set.} We select hyper-parameters $\gamma$, $\alpha$ from $\{0.4, 0.6, 0.8, 2, 3, 4, 5\}$.

\paragraph{Results:} Table~\ref{tab:transfer_results} shows the results of the debiased models and baseline with BERT. As shown in prior work~\citep{belinkov-etal-2019-dont}, the MNLI datasets have very similar biases to SNLI, which the models are trained on, so we 
do not expect any improvement in the relative performance of our models and the
baseline for MNLI and MNLI-M. On all the remaining datasets, our proposed models perform better than the baseline, showing a substantial improvement in generalization by using our debasing techniques. We additionally compare with \citet{belinkov-etal-2019-dont} in Appendix \ref{app:transfer} and show that our methods substantially surpass their results.

\begin{table}
    \centering
    \begin{tabular}{llllllllll}
        \toprule 
       \textbf{Data}  & \textbf{CE} & \textbf{DFL} & \bm{$\Delta$} &\textbf{PoE} & \bm{$\Delta$} \\  
        \toprule  
        SICK & 57.05  & 57.91 & +0.9 &57.28 & +0.2\\
        ADD1 & 87.34  &88.89 & +1.5 & 87.86 & +0.5\\
        DPR &  49.50 & 50.68 & +1.2 &50.14 &+0.6\\
        SPR& 59.85   &61.41 & +1.6&62.45 &+2.6\\
        FN+ &  53.16  &54.77 &+1.6 &53.51 & +0.4\\
        JOCI & 50.06  &51.13 &+1.1 &50.85  & +0.8\\
        MPE &  69.50  &70.2 &+0.7 & 70.1 & +0.6\\
        SCITAIL&67.64 & 69.33 & +1.7&71.40 &+3.8\\
        GLUE & 54.08  & 54.80 & +0.7  & 54.71 & +0.6\\
        QQP &  67.78  & 69.28 &+1.5 &68.61 & +0.8\\
        MNLI & 74.40& 73.58 &-0.8& 73.61& -0.8\\
        MNLI-M &73.98  & 74.0 & 0.0  & 73.49 & -0.5\\
    \bottomrule
    \end{tabular}
 \caption{Accuracy results of models with BERT transferring to new target datasets. All models are trained on SNLI and tested on the target datasets. \bm{$\Delta$} are absolute differences between our methods and the CE loss baseline.}
    \label{tab:transfer_results}
\end{table}

 \begin{figure}[H]
    \centering 
    \includegraphics[width=0.4\textwidth]{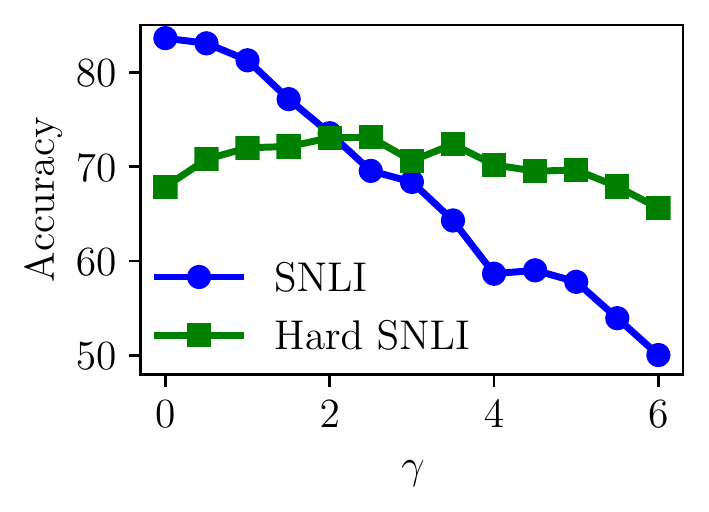} 
    \caption{Accuracy of InferSent model trained with DFL, on the SNLI test and SNLI hard sets for different $\gamma$.} 
    \label{fig:debiased_focal_loss}
\end{figure}

\section{Discussion}
\paragraph{Analysis of Debiased Focal Loss:} As expected, improving the out-of-domain performance could come at the expense of decreased in-domain performance since the removed biases are useful for performing the in-domain task. This  happens especially for DFL, in which there is a trade-off between in-domain and out-of-domain performance that depends on the parameter $\gamma$, and when the baseline model is not very powerful like InferSent. 
To understand the impact of $\gamma$ in DFL, we train an InferSent model using DFL for different values of $\gamma$ on the SNLI dataset and evaluate its performance on SNLI test and SNLI hard sets. As illustrated in Figure~\ref{fig:debiased_focal_loss}, increasing $\gamma$ increases debiasing and thus hurts in-domain accuracy on SNLI, but out-of-domain accuracy on the SNLI hard set is increased within a wide range of values (see a similar plot for BERT in Appendix ~\ref{app:discussion}).

\begin{figure}[H]
 \centering 
    \includegraphics[width=0.46\textwidth]{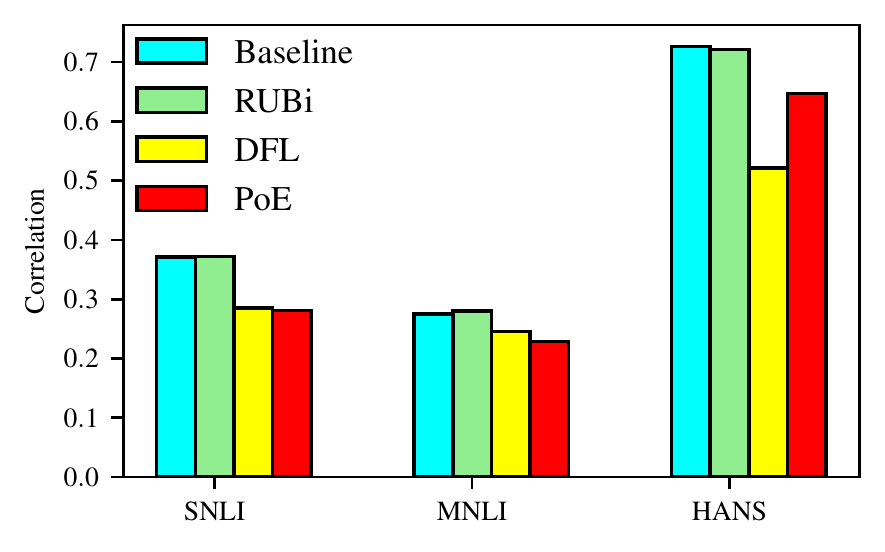}
\caption{Pearson correlation between the element-wise cross-entropy loss of the debiasing models and the bias-only model trained on each dataset.}
    \label{fig:correlation}
\end{figure}

\paragraph{Correlation Analysis:} In contrast to~\citet{belinkov-etal-2019-dont}, who encourage only the encoder to not capture the unwanted biases, our learning strategies influence the parameters of the full model to reduce the reliance on unwanted patterns more effectively. To test this assumption, in Figure~\ref{fig:correlation}, we report the correlation between the element-wise loss of the debiased models and the loss of a bias-only model on the considered datasets.

The results show that compared to the baselines, our debiasing methods, DFL and PoE,  reduce the correlation to the bias-only model, confirming that our models are effective at reducing biases. Interestingly, on MNLI, PoE has less correlation with the bias-only model than DFL and also has better performance on the unbiased split of this dataset. On the other hand, on the HANS dataset, DFL loss is less correlated with the bias-only model than PoE and also obtains higher performance on the HANS dataset.

\section{Conclusion}
We propose two novel techniques, product-of-experts and debiased focal loss, to reduce biases learned by neural models, which are applicable whenever one can specify the biases in the form of one or more bias-only models. 
The bias-only models are designed to leverage biases and shortcuts in the datasets. Our debiasing strategies then 
work by adjusting the cross-entropy loss based on the  performance of these bias-only
models, to focus learning on the hard examples and down-weight the importance
of the biased examples. Additionally, we extend our methods to combat multiple bias patterns simultaneously. Our proposed  debiasing techniques are model agnostic, simple, and highly effective. 
Extensive experiments show that our methods substantially improve the model robustness to domain-shift, including 
 9.8 points gain on FEVER symmetric test set, 7.4 on HANS dataset, and 4.8 points on SNLI hard set. Furthermore, we show that our debiasing techniques result in better generalization  to other NLI datasets. 
Future work may include developing debiasing strategies that  do not require prior knowledge of bias patterns and can automatically identify them.

\subsubsection*{Acknowledgments}
We would like to thank Daniel Andor and Suraj Srinivas for their helpful comments. 
We additionally would like to thank the authors of \citet{schuster2019towards, cadene2019rubi, mccoy-etal-2019-right, belinkov-etal-2019-dont} for their support to reproduce their results.
This research was supported by the Swiss National Science Foundation under the project Learning
Representations of Abstraction for Opinion Summarization (LAOS), grant number ``FNS-30216''.
Y.B.\ was supported by the Harvard Mind, Brain, and Behavior Initiative. 

\bibliography{acl2020} 

\begin{thebibliography}{39}
\expandafter\ifx\csname natexlab\endcsname\relax\def\natexlab#1{#1}\fi

\bibitem[{Belinkov et~al.(2019{\natexlab{a}})Belinkov, Poliak, Shieber,
  Van~Durme, and Rush}]{belinkov-etal-2019-dont}
Yonatan Belinkov, Adam Poliak, Stuart Shieber, Benjamin Van~Durme, and
  Alexander Rush. 2019{\natexlab{a}}.
\newblock \href {https://doi.org/10.18653/v1/P19-1084} {Don{'}t take the
  premise for granted: Mitigating artifacts in natural language inference}.
\newblock In \emph{Proceedings of the 57th Annual Meeting of the Association
  for Computational Linguistics}, pages 877--891, Florence, Italy. Association
  for Computational Linguistics.

\bibitem[{Belinkov et~al.(2019{\natexlab{b}})Belinkov, Poliak, Shieber,
  Van~Durme, and Rush}]{belinkov2019adversarial}
Yonatan Belinkov, Adam Poliak, Stuart~M Shieber, Benjamin Van~Durme, and
  Alexander~M Rush. 2019{\natexlab{b}}.
\newblock On adversarial removal of hypothesis-only bias in natural language
  inference.
\newblock In \emph{Proceedings of the 8th Joint Conference on Lexical and
  Computational Semantics}.

\bibitem[{Bowman et~al.(2015)Bowman, Angeli, Potts, and
  Manning}]{bowman2015large}
Samuel~R Bowman, Gabor Angeli, Christopher Potts, and Christopher~D Manning.
  2015.
\newblock A large annotated corpus for learning natural language inference.
\newblock In \emph{Proceedings of the 2015 Conference on Empirical Methods in
  Natural Language Processing}.

\bibitem[{Cadene et~al.(2019)Cadene, Dancette, Ben-younes, Cord, and
  Parikh}]{cadene2019rubi}
Remi Cadene, Corentin Dancette, Hedi Ben-younes, Matthieu Cord, and Devi
  Parikh. 2019.
\newblock Rubi: Reducing unimodal biases in visual question answering.
\newblock In \emph{Advances in neural information processing systems}.

\bibitem[{Chen et~al.(2017)Chen, Zhu, Ling, Wei, Jiang, and
  Inkpen}]{chen2017enhanced}
Qian Chen, Xiaodan Zhu, Zhen-Hua Ling, Si~Wei, Hui Jiang, and Diana Inkpen.
  2017.
\newblock Enhanced lstm for natural language inference.
\newblock In \emph{Proceedings of the 55th Annual Meeting of the Association
  for Computational Linguistics}.

\bibitem[{Clark et~al.(2019)Clark, Yatskar, and Zettlemoyer}]{clark2019dont}
Christopher Clark, Mark Yatskar, and Luke Zettlemoyer. 2019.
\newblock Don't take the easy way out: Ensemble based methods for avoiding
  known dataset biases.
\newblock In \emph{Proceedings of the 2019 Conference on Empirical Methods in
  Natural Language Processing}.

\bibitem[{Conneau et~al.(2017)Conneau, Kiela, Schwenk, Barrault, and
  Bordes}]{conneau2017supervised}
Alexis Conneau, Douwe Kiela, Holger Schwenk, Lo{\"\i}c Barrault, and Antoine
  Bordes. 2017.
\newblock Supervised learning of universal sentence representations from
  natural language inference data.
\newblock In \emph{Proceedings of the 2017 Conference on Empirical Methods in
  Natural Language Processing}.

\bibitem[{Dagan et~al.(2006)Dagan, Glickman, and Magnini}]{dagan2006pascal}
Ido Dagan, Oren Glickman, and Bernardo Magnini. 2006.
\newblock The pascal recognising textual entailment challenge.
\newblock \emph{Machine Learning Challenges. Evaluating Predictive Uncertainty,
  Visual Object Classification, and Recognising Tectual Entailment}.

\bibitem[{Devlin et~al.(2019)Devlin, Chang, Lee, and
  Toutanova}]{devlin2018bert}
Jacob Devlin, Ming-Wei Chang, Kenton Lee, and Kristina Toutanova. 2019.
\newblock Bert: Pre-training of deep bidirectional transformers for language
  understanding.
\newblock In \emph{Proceedings of the 2019 Conference of the North American
  Chapter of the Association for Computational Linguistics: Human Language
  Technologies}.

\bibitem[{Gong et~al.(2017)Gong, Luo, and Zhang}]{gong2017natural}
Yichen Gong, Heng Luo, and Jian Zhang. 2017.
\newblock Natural language inference over interaction space.
\newblock \emph{In International Conference on Learning Representations}.

\bibitem[{Grand and Belinkov(2019)}]{grand2019adversarial}
Gabriel Grand and Yonatan Belinkov. 2019.
\newblock Adversarial regularization for visual question answering: Strengths,
  shortcomings, and side effects.
\newblock In \emph{Proceedings of the Second Workshop on Shortcomings in Vision
  and Language}, pages 1--13.

\bibitem[{Gururangan et~al.(2018)Gururangan, Swayamdipta, Levy, Schwartz,
  Bowman, and Smith}]{gururangan2018annotation}
Suchin Gururangan, Swabha Swayamdipta, Omer Levy, Roy Schwartz, Samuel Bowman,
  and Noah~A Smith. 2018.
\newblock Annotation artifacts in natural language inference data.
\newblock In \emph{Proceedings of the 2018 Conference of the North American
  Chapter of the Association for Computational Linguistics: Human Language
  Technologies}.

\bibitem[{He et~al.(2019)He, Zha, and Wang}]{he-etal-2019-unlearn}
He~He, Sheng Zha, and Haohan Wang. 2019.
\newblock \href {https://doi.org/10.18653/v1/D19-6115} {Unlearn dataset bias in
  natural language inference by fitting the residual}.
\newblock In \emph{Proceedings of the 2nd Workshop on Deep Learning Approaches
  for Low-Resource NLP (DeepLo 2019)}, pages 132--142, Hong Kong, China.
  Association for Computational Linguistics.

\bibitem[{Hinton(2002)}]{hinton2002training}
Geoffrey~E Hinton. 2002.
\newblock Training products of experts by minimizing contrastive divergence.
\newblock \emph{Neural computation}.

\bibitem[{Jia and Liang(2017)}]{jia2017adversarial}
Robin Jia and Percy Liang. 2017.
\newblock Adversarial examples for evaluating reading comprehension systems.
\newblock In \emph{Proceedings of the 2017 Conference on Empirical Methods in
  Natural Language Processing}.

\bibitem[{Joulin et~al.(2017)Joulin, Grave, Bojanowski, and
  Mikolov}]{joulin2017bag}
Armand Joulin, Edouard Grave, Piotr Bojanowski, and Tomas Mikolov. 2017.
\newblock Bag of tricks for efficient text classification.
\newblock In \emph{Proceedings of the 15th Conference of the European Chapter
  of the Association for Computational Linguistics}.

\bibitem[{Kaushik and Lipton(2018)}]{kaushik2018much}
Divyansh Kaushik and Zachary~C Lipton. 2018.
\newblock How much reading does reading comprehension require? a critical
  investigation of popular benchmarks.
\newblock In \emph{Proceedings of the 2018 Conference on Empirical Methods in
  Natural Language Processing}.

\bibitem[{Khot et~al.(2018)Khot, Sabharwal, and Clark}]{khot2018scitail}
Tushar Khot, Ashish Sabharwal, and Peter Clark. 2018.
\newblock Scitail: A textual entailment dataset from science question
  answering.
\newblock In \emph{Thirty-Second AAAI Conference on Artificial Intelligence}.

\bibitem[{Lai et~al.(2017)Lai, Bisk, and Hockenmaier}]{lai2017natural}
Alice Lai, Yonatan Bisk, and Julia Hockenmaier. 2017.
\newblock Natural language inference from multiple premises.
\newblock In \emph{Proceedings of the Eighth International Joint Conference on
  Natural Language Processing}.

\bibitem[{Lin et~al.(2017)Lin, Goyal, Girshick, He, and
  Doll{\'a}r}]{lin2017focal}
Tsung-Yi Lin, Priya Goyal, Ross Girshick, Kaiming He, and Piotr Doll{\'a}r.
  2017.
\newblock Focal loss for dense object detection.
\newblock In \emph{Proceedings of the IEEE international conference on computer
  vision}.

\bibitem[{Marelli et~al.(2014)Marelli, Menini, Baroni, Bentivogli, Bernardi,
  and Zamparelli}]{MARELLI14.363}
Marco Marelli, Stefano Menini, Marco Baroni, Luisa Bentivogli, Raffaella
  Bernardi, and Roberto Zamparelli. 2014.
\newblock A sick cure for the evaluation of compositional distributional
  semantic models.
\newblock In \emph{Proceedings of the Ninth International Conference on
  Language Resources and Evaluation (LREC'14)}, Reykjavik, Iceland. European
  Language Resources Association (ELRA).

\bibitem[{McCoy et~al.(2019{\natexlab{a}})McCoy, Min, and
  Linzen}]{mccoy2019berts}
R~Thomas McCoy, Junghyun Min, and Tal Linzen. 2019{\natexlab{a}}.
\newblock Berts of a feather do not generalize together: Large variability in
  generalization across models with similar test set performance.
\newblock \emph{arXiv preprint arXiv:1911.02969}.

\bibitem[{McCoy et~al.(2019{\natexlab{b}})McCoy, Pavlick, and
  Linzen}]{mccoy-etal-2019-right}
Tom McCoy, Ellie Pavlick, and Tal Linzen. 2019{\natexlab{b}}.
\newblock \href {https://doi.org/10.18653/v1/P19-1334} {Right for the wrong
  reasons: Diagnosing syntactic heuristics in natural language inference}.
\newblock In \emph{Proceedings of the 57th Annual Meeting of the Association
  for Computational Linguistics}, pages 3428--3448, Florence, Italy.
  Association for Computational Linguistics.

\bibitem[{Mou et~al.(2016)Mou, Men, Li, Xu, Zhang, Yan, and
  Jin}]{mou2016natural}
Lili Mou, Rui Men, Ge~Li, Yan Xu, Lu~Zhang, Rui Yan, and Zhi Jin. 2016.
\newblock Natural language inference by tree-based convolution and heuristic
  matching.
\newblock In \emph{Proceedings of the 54th Annual Meeting of the Association
  for Computational Linguistics}.

\bibitem[{Pavlick and Callison-Burch(2016)}]{pavlick2016most}
Ellie Pavlick and Chris Callison-Burch. 2016.
\newblock Most" babies" are" little" and most" problems" are" huge":
  Compositional entailment in adjective-nouns.
\newblock In \emph{Proceedings of the 54th Annual Meeting of the Association
  for Computational Linguistics}.

\bibitem[{Pavlick et~al.(2015)Pavlick, Wolfe, Rastogi, Callison-Burch, Dredze,
  and Van~Durme}]{pavlick2015framenet+}
Ellie Pavlick, Travis Wolfe, Pushpendre Rastogi, Chris Callison-Burch, Mark
  Dredze, and Benjamin Van~Durme. 2015.
\newblock Framenet+: Fast paraphrastic tripling of framenet.
\newblock In \emph{Proceedings of the 53rd Annual Meeting of the Association
  for Computational Linguistics and the 7th International Joint Conference on
  Natural Language Processing (Volume 2: Short Papers)}, volume~2, pages
  408--413.

\bibitem[{Poliak et~al.(2018)Poliak, Naradowsky, Haldar, Rudinger, and
  Van~Durme}]{poliak2018hypothesis}
Adam Poliak, Jason Naradowsky, Aparajita Haldar, Rachel Rudinger, and Benjamin
  Van~Durme. 2018.
\newblock Hypothesis only baselines in natural language inference.
\newblock In \emph{Proceedings of the Seventh Joint Conference on Lexical and
  Computational Semantics}.

\bibitem[{Radford et~al.(2018)Radford, Narasimhan, Salimans, and
  Sutskever}]{radford2018improving}
Alec Radford, Karthik Narasimhan, Tim Salimans, and Ilya Sutskever. 2018.
\newblock \href
  {https://www.cs.ubc.ca/~amuham01/LING530/papers/radford2018improving.pdf}
  {Improving language understanding by generative pre-training}.

\bibitem[{Rahman and Ng(2012)}]{rahman2012resolving}
Altaf Rahman and Vincent Ng. 2012.
\newblock Resolving complex cases of definite pronouns: the winograd schema
  challenge.
\newblock In \emph{Proceedings of the 2012 Joint Conference on Empirical
  Methods in Natural Language Processing and Computational Natural Language
  Learning}.

\bibitem[{Ramakrishnan et~al.(2018)Ramakrishnan, Agrawal, and
  Lee}]{ramakrishnan2018overcoming}
Sainandan Ramakrishnan, Aishwarya Agrawal, and Stefan Lee. 2018.
\newblock Overcoming language priors in visual question answering with
  adversarial regularization.
\newblock In \emph{Advances in Neural Information Processing Systems}, pages
  1541--1551.

\bibitem[{Reisinger et~al.(2015)Reisinger, Rudinger, Ferraro, Harman, Rawlins,
  and Van~Durme}]{reisinger2015semantic}
Drew Reisinger, Rachel Rudinger, Francis Ferraro, Craig Harman, Kyle Rawlins,
  and Benjamin Van~Durme. 2015.
\newblock Semantic proto-roles.
\newblock \emph{Transactions of the Association for Computational Linguistics}.

\bibitem[{Schuster et~al.(2019)Schuster, J~Shah, Jie Serene~Yeo, Filizzola,
  Santus, and Barzilay}]{schuster2019towards}
Tal Schuster, Darsh J~Shah, Yun Jie Serene~Yeo, Daniel Filizzola, Enrico
  Santus, and Regina Barzilay. 2019.
\newblock Towards debiasing fact verification models.
\newblock In \emph{Proceedings of the 2019 Conference on Empirical Methods in
  Natural Language Processing}.

\bibitem[{Sharma et~al.(2018)Sharma, Allen, Bakhshandeh, and
  Mostafazadeh}]{sharma2018tackling}
Rishi Sharma, James Allen, Omid Bakhshandeh, and Nasrin Mostafazadeh. 2018.
\newblock Tackling the story ending biases in the story cloze test.
\newblock In \emph{Proceedings of the 56th Annual Meeting of the Association
  for Computational Linguistics}.

\bibitem[{Wang et~al.(2019)Wang, Singh, Michael, Hill, Levy, and
  Bowman}]{wang2018glue}
Alex Wang, Amapreet Singh, Julian Michael, Felix Hill, Omer Levy, and Samuel~R.
  Bowman. 2019.
\newblock {GLUE}: A multi-task benchmark and analysis platform for natural
  language understanding.
\newblock In \emph{International Conference on Learning Representations}.

\bibitem[{Wang et~al.(2017)Wang, Hamza, and Florian}]{wang2017bilateral}
Zhiguo Wang, Wael Hamza, and Radu Florian. 2017.
\newblock Bilateral multi-perspective matching for natural language sentences.
\newblock In \emph{Proceedings of the 26th International Joint Conference on
  Artificial Intelligence}.

\bibitem[{White et~al.(2017)White, Rastogi, Duh, and
  Van~Durme}]{white2017inference}
Aaron~Steven White, Pushpendre Rastogi, Kevin Duh, and Benjamin Van~Durme.
  2017.
\newblock Inference is everything: Recasting semantic resources into a unified
  evaluation framework.
\newblock In \emph{Proceedings of the Eighth International Joint Conference on
  Natural Language Processing}.

\bibitem[{Williams et~al.(2018)Williams, Nangia, and
  Bowman}]{williams2018broad}
Adina Williams, Nikita Nangia, and Samuel Bowman. 2018.
\newblock A broad-coverage challenge corpus for sentence understanding through
  inference.
\newblock In \emph{Proceedings of the 2018 Conference of the North American
  Chapter of the Association for Computational Linguistics: Human Language
  Technologies}.

\bibitem[{Wolf et~al.(2019)Wolf, Debut, Sanh, Chaumond, Delangue, Moi, Cistac,
  Rault, Louf, Funtowicz, and Brew}]{Wolf2019HuggingFacesTS}
Thomas Wolf, Lysandre Debut, Victor Sanh, Julien Chaumond, Clement Delangue,
  Anthony Moi, Pierric Cistac, Tim Rault, R'emi Louf, Morgan Funtowicz, and
  Jamie Brew. 2019.
\newblock Huggingface's transformers: State-of-the-art natural language
  processing.
\newblock \emph{ArXiv}, abs/1910.03771.

\bibitem[{Zhang et~al.(2017)Zhang, Rudinger, Duh, and
  Van~Durme}]{zhang2017ordinal}
Sheng Zhang, Rachel Rudinger, Kevin Duh, and Benjamin Van~Durme. 2017.
\newblock Ordinal common-sense inference.
\newblock \emph{Transactions of the Association for Computational Linguistics}.

\end{thebibliography}
\bibliographystyle{acl_natbib}

\clearpage

\appendix
\section{Fact Verification} \label{sec:appendix_verification}
\paragraph{Base model:}  We fine-tune all models using BERT for 3 epochs and use the default parameters and default learning rate of $2\mathrm{e}{-5}$. 

\paragraph{Bias-only model:}  Our bias-only classifier is a shallow nonlinear classifier with 768, 384, 192 hidden units with Tanh nonlinearity.

\section{Natural Language Inference} \label{appendix:entailment}
\paragraph{Base model:}  InferSent uses a separate BiLSTM encoder to learn sentence representations for premise and hypothesis.
It then combines these embeddings following~\citet{mou2016natural} and feeds them to the default nonlinear classifier.
With InferSent we train all models for 20 epochs as default without using early-stopping. We use the default hyper-parameters and following~\citet{wang2018glue}, we set the BiLSTM dimension to 512.
We use the default nonlinear classifier with 512 and 512 hidden neurons with Tanh nonlinearity. With BERT, we finetune all models for 3 epochs.

\paragraph{Bias-only model:}  For debiasing models using BERT, we use the 
same shallow nonlinear classifier explained in Appendix~\ref{sec:appendix_verification}, and for the ones using InferSent, we use 
a shallow linear classifier with 512 and 512 hidden units.

\paragraph{Results:} Table~\ref{tab:mnli_matched_hard_test_results_all} shows results on the MNLI matched development and hard test sets.

\begin{table}[H]
    \centering
      \begin{center}
    \resizebox{0.48\textwidth}{!}{
    \begin{tabular}{lllllllllllllll} 
        \toprule 
         & \multicolumn{3}{c}{\textbf{BERT}} & \multicolumn{3}{c}{\textbf{InferSent}} \\ 
         \cmidrule(r){2-4} \cmidrule(l){5-7} 
        {\vspace{-0.75em} \textbf{Loss}} &  \multicolumn{2}{c}{} & \multicolumn{2}{c}{}\\ 
                 & \textbf{MNLI} & \textbf{Hard} & \bm{$\Delta$} &\textbf{MNLI} & \textbf{Hard} & \bm{$\Delta$} \\  
        \toprule 
         \multicolumn{7}{c}{\textbf{Development set results}}\\
         \toprule  
        CE & 84.41 &   \textbf{76.56}&&  69.97 &  \textbf{57.03}& \\
        RUBi & 84.48 &    77.13 & +0.6 & 70.51 &  57.97&+0.9\\
        \midrule 
        DFL  &83.72 &   77.37 &+0.8& 60.78 & 57.88&+0.9\\ 
        PoE &84.58 &   \textbf{78.02} &\textbf{+1.5}& 66.02 &  \textbf{59.37} &\textbf{+2.3}\\ 
    \bottomrule
    \multicolumn{7}{c}{\textbf{Test set results}} \\
    \toprule
    None & 84.11 & 75.88  & & ---&--- &---  \\
    PoE & 84.11 & \textbf{76.81}  &\textbf{+0.9} & ---&---&--- \\
    \bottomrule
    \end{tabular}}
    \end{center}
 \caption{Results on the MNLI matched benchmark and MNLI matched hard set. $\bm{\Delta}$ are absolute differences with CE loss.}    \label{tab:mnli_matched_hard_test_results_all}
\end{table}

\section{Syntactic Bias in NLI} \label{appendix:hans}
\paragraph{Base model:}  We finetune all models for 3 epochs. 

\paragraph{Bias-only model:} We use a nonlinear classifier with 6 and 6 hidden units with Tanh nonlinearity. 

\paragraph{Results:} Table~\ref{tab:hans-results-individual} shows the performance for each label (entailment and non\_entailment) on individual heuristics of the HANS dataset.

\begin{table}[H]
\centering 
\resizebox{0.48\textwidth}{!}{
\begin{tabular}{lccc} \toprule
 \multirow{2}{*}{\textbf{Loss}} & \multicolumn{3}{c}{\textbf{HANS}} \\  
 \cmidrule(r){2-4}
 &  \textbf{Constituent} & \textbf{Lexical} & \textbf{Subsequence} \\
\toprule 
\multicolumn{4}{c}{\textbf{gold label: Entailment}} \\ 
\toprule 
CE   & 98.98$\pm$0.6 & 96.41$\pm$0.8  & 99.72$\pm$0.1 \\
RUBi & 99.22$\pm$0.3 & 95.59$\pm$0.8 & 99.50$\pm$0.3\\ 
\midrule 
DFL & 90.90$\pm$4.3 & 84.78$\pm$5.0 & 94.33$\pm$4.9 \\
PoE & 97.24$\pm$1.9 & 92.16$\pm$0.9 & 98.58$\pm$0.5\\ 
\toprule  
\multicolumn{4}{c}{\textbf{gold label: Non-entailment}} \\ 
\toprule 
CE   &  20.12$\pm$5.8 & 48.86$\pm$5.7 &7.18$\pm$0.7   \\
RUBi & 21.89$\pm$7.0 & 46.82$\pm$12.5 & 7.58$\pm$2.3\\
\midrule 
DFL  & 50.20$\pm$9.2 &71.06$\pm$3.1 &24.28$\pm$4.4\\
PoE &36.08$\pm$5.1 &59.18$\pm$8.0  &14.63$\pm$3.0 \\
\bottomrule
\end{tabular}}
\caption{Accuracy for each label (entailment or non-entailment) on individual heuristics of HANS.} 
\label{tab:hans-results-individual}
\end{table}

\begin{table*}
\centering 
\begin{tabular}{llllllllll} 
\toprule
\textbf{Data}  &   \textbf{CE} & \textbf{DFL} &  $\bf{\Delta} \%$ &    \textbf{PoE} & $\bf{\Delta} \%$ &  \textbf{M1} &  $\bf{\Delta} \%$ &  \textbf{M2} &  $\bf{\Delta} \%$ \\ 
\midrule
SICK         &  54.09 &  55.00 &       1.68 &  55.79 &       3.14 &   49.77 &        -7.99 &   49.77 &        -7.99 \\ 
ADD1      &  75.19 &  78.29 &       4.12 &  77.00 &       2.41 &   67.44 &       -10.31 &   67.44 &       -10.31 \\
DPR            &  49.95 &  50.59 &       1.28 &  49.95 &       0.00 &   50.87 &         1.84 &   50.87 &         1.84\\
SPR           &  41.31 &  47.95 &      16.07 &  50.50 &      22.25 &   51.51 &        24.69 &   51.51 &        24.69 \\
FN+         &  48.65 &  49.58 &       1.91 &  49.35 &       1.44 &   53.23 &         9.41 &   53.23 &         9.41 \\
JOCI           &  46.47 &  46.48 &       0.02 &  47.53 &       2.28 &   44.83 &        -3.53 &   44.83 &        -3.53 \\
MPE            &  60.60 &  60.70 &       0.17 &  61.80 &       1.98 &   56.40 &        -6.93 &   56.40 &        -6.93 \\
SCITAIL        &  64.25 &  65.19 &       1.46 &  63.17 &      -1.68 &   56.40 &       -12.22 &   56.40 &       -12.22 \\
GLUE &  48.73 &  46.83 &      -3.90 &  49.09 &       0.74 &   43.93 &        -9.85 &   43.93 &        -9.85 \\ 
QQP            &  61.80 &  66.24 &       7.18 &  66.36 &       7.38 &   62.46 &         1.07 &   62.46 &         1.07 \\
MNLI    &  56.99 &  56.70 &      -0.51 &  56.59 &      -0.70 &   51.72 &        -9.25 &   51.72 &        -9.25 \\
MNLI-M &  57.01 &  57.75 &       1.30 &  57.84 &       1.46 &   53.99 &        -5.30 &   53.99 &        -5.30 \\ 
\midrule 
Average           &  --- &  --- &       2.57 &  --- &       3.39 &   --- &        -2.36 &   --- &        -2.36\\
\bottomrule
\end{tabular}
 \caption{Accuracy results of models with InferSent transferring to new target datasets. All models are trained on SNLI and tested on the target datasets. M1 and M2 are our re-implementation of ~\citet{belinkov-etal-2019-dont}.
 $\bf{\Delta}$ are relative differences in percentage  with respect to CE loss.}. \vspace{-1em}
 \label{tab:infersent_comparisons}
\end{table*}

\section{Transfer Performance} \label{app:transfer}
\paragraph{Mapping:} We train all models on SNLI and evaluate their performance on other target datasets. SNLI contains three labels,  contradiction,  neutral, and  entailment. Some of the datasets we consider contain only two labels.  In the case of labels \emph{entailed} and \emph{not-entailed}, as in DPR, we map contradiction and  neutral to the not-entailed class.  In the case of labels \emph{entailment} and \emph{neutral}, as in SciTail, we map contradiction to neutral.

\paragraph{Comparison with \citet{belinkov-etal-2019-dont}:}
We modified the implementations of ~\citet{belinkov-etal-2019-dont} and corrected some implementation issues in the InferSent baseline~\cite{conneau2017supervised}. Compared to the original InferSent implementation, the main differences in our implementation include: (a) We incorporated the fixes suggested for the bugs in the implementation of mean/max-pooling over BiLSTM in the InferSent baseline\footnote{\url{https://github.com/facebookresearch/InferSent/issues/51}} (b).  We additionally observed that the aggregation of losses over each batch was computed with the average instead of the intended summation and we corrected it.\footnote{The same observation is reported in \url{https://github.com/facebookresearch/InferSent/pull/107}.} (c) We followed the implementation of InferSent and we removed out-of-vocabulary (OOV) words from the sentence representation, while~\citeauthor{belinkov-etal-2019-dont} keep them by introducing an OOV token.
We additionally observed during the pre-processing of some of the target datasets in the implementation of ~\citeauthor{belinkov-etal-2019-dont}, some of the samples are not considered due to the preprocessing issues. We fix the pre-processing issues and evaluate our models and our reimplementations of ~\citet{belinkov-etal-2019-dont} on the same corpora. We set the BiLSTM dimension to 512 across all models. Note that~\citeauthor{belinkov-etal-2019-dont} use BiLSTM dimension of 2048, and due to the mentioned differences in implementations and datasets, the results reported in~\citet{belinkov-etal-2019-dont} are not comparable. However, we still on average surpass their reported results substantially. Our reimplementations and scripts to reproduce the results are publicly available in \url{https://github.com/rabeehk/robust-nli-fixed}.

As used in prior work to adjust the learning-rate of the bias-only
and baseline models~\citep{belinkov-etal-2019-dont}, we introduce a hyperparameter $\beta$ for the bias-only model to modulate the loss of the bias-only model in ensembling. We sweep hyper-parameters $\gamma, \alpha$ over $\{0.02, 0.05, 0.1, 0.6, 2.0, 4.0, 5.0\}$ and $\beta$ over $\{0.05, 0.2, 0.4, 0.8, 1.0\}$. Table \ref{tab:infersent_comparisons}
shows the results of our debiasing models (DFL, PoE), our re-implementations of proposed methods in~\citet{belinkov-etal-2019-dont} (M1, M2), and the baseline with InferSent (CE). The DFL model outperforms the baseline in 10 out of 12 datasets, while the PoE model outperforms the baseline in 9 datasets and does equally well on the DPR dataset.  As shown in prior work~\citep{belinkov-etal-2019-dont},  the MNLI dataset has very similar biases to SNLI, which the models are trained on, so we do not expect any improvement in  the  relative  performance  of  our  models  and the  baseline  for  MNLI  dataset. Interestingly, our methods obtain improvement on MNLI-M, in which the test data differs from  training distribution. Our proposed debiasing methods, PoE and DFL, are highly effective, boosting the relative generalization performance of the baseline by 3.39\% and 2.57\% respectively, significantly surpassing the prior work of~\citet{belinkov-etal-2019-dont}. Compared to M1 and M2, our methods outperform them on 9 datasets, while they do better on two datasets of SPR and FN+, and slightly better on the DPR dataset. However, note that DPR is a very small dataset and all models perform close to random-chance on this dataset.

\section{Analysis of Debiased Focal Loss} \label{app:discussion}
\begin{figure}[H]
    \centering 
    \includegraphics[width=0.42\textwidth]{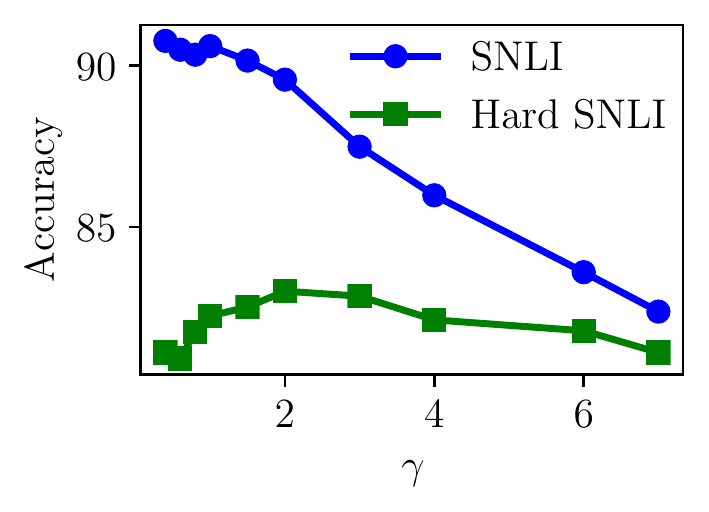} 
    \caption{Accuracy of the BERT model trained with DFL, on SNLI and SNLI hard sets for different $\gamma$. }
    \label{fig:debiased_focal_loss_bert}\vspace{-1.5em}
\end{figure}

Figure ~\ref{fig:debiased_focal_loss_bert} shows the impact of $\gamma$ on BERT trained with DFL.

\end{document}